\pdfoutput=1
\documentclass[10pt,twocolumn,letterpaper]{article}

\usepackage{cvpr}              
\usepackage{float}
\usepackage{algorithm}
\usepackage{algorithmic}
\usepackage{xcolor}
\usepackage{url}
\definecolor{cvprblue}{rgb}{0.21,0.49,0.74}
\usepackage[pagebackref,breaklinks,colorlinks,allcolors=cvprblue]{hyperref}

\def\paperID{11000} 
\def\confName{CVPR}
\def\confYear{2025}

\title{Encapsulated Composition of Text-to-Image and Text-to-Video Models for High-Quality Video Synthesis}

\author{Tongtong Su$^{1,2}$, Chengyu Wang$^{2}$\footnotemark[1], Bingyan Liu$^{3,2}$, Jun Huang$^2$, Dongming Lu$^{1}\footnotemark[1]$\\
$^1$ Zhejiang University,
$^2$ Alibaba Cloud Computing,\\
$^3$ South China University of Technology\\
{
\tt\small \{sutongtong,ldm\}@zju.edu.cn, 
    }\\
    {
\tt\small \{chengyu.wcy,huangjun.hj\}@alibaba-inc.com,
    }
    {
\tt\small eeliubingyan@mail.scut.edu.cn
    }
}

\begin{document}

\twocolumn[{
\maketitle
\begin{figure}[H]
\hsize=\textwidth 
\centering
\includegraphics[width=.90\textwidth]{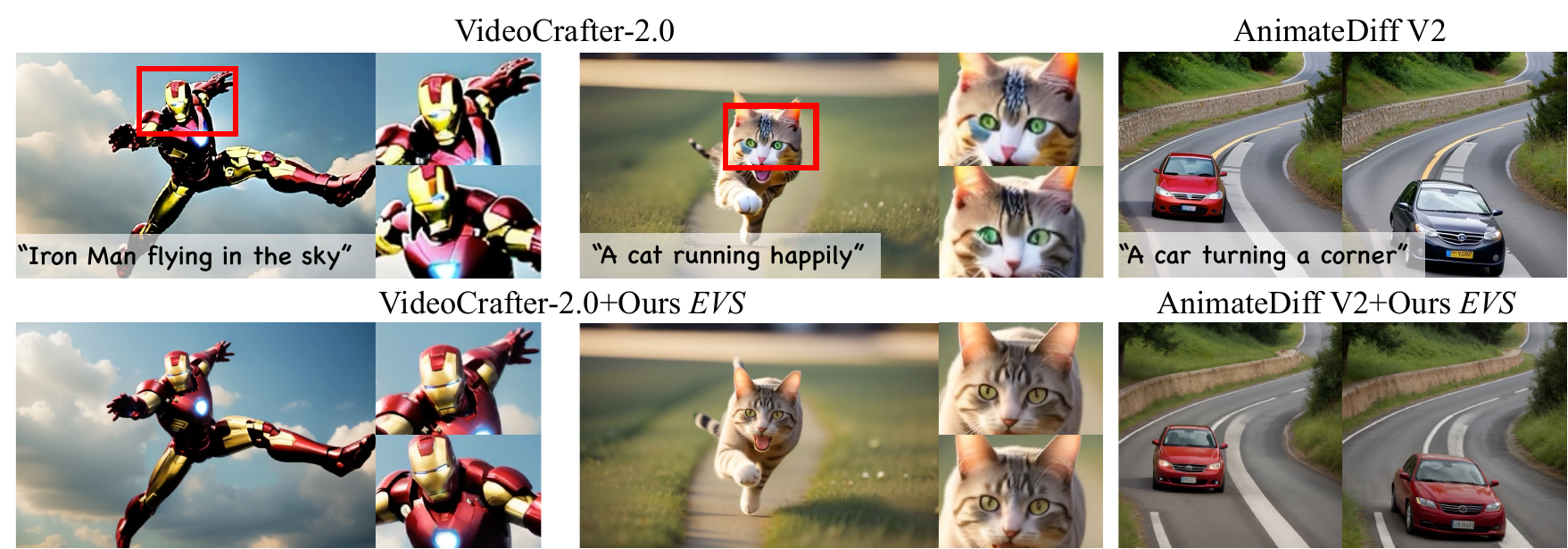}
\caption{Comparison of videos generated by T2V models w/ and w/o~\emph{EVS}. Our method significantly improves imaging quality compared to videos solely generated by VideoCrafter-2.0 (see the Iron Man head and cat eyes), and frame consistency compared to AnimateDiff-V2 (see the stability of car color across frames).}
\label{fig:intro_demo}
\end{figure}
 }]

\renewcommand{\thefootnote}{\fnsymbol{footnote}}
\footnotetext[1]{Co-corresponding authors.}
\renewcommand{\thefootnote}{\arabic{footnote}}

\begin{abstract}

In recent years, large text-to-video (T2V) synthesis models have garnered considerable attention for their abilities to generate videos from textual descriptions. However, achieving both high imaging quality and effective motion representation remains a significant challenge for these T2V models. Existing approaches often adapt pre-trained text-to-image (T2I) models to refine video frames, leading to issues such as flickering and artifacts due to inconsistencies across frames. In this paper, we introduce \emph{EVS}, a training-free \underline{E}ncapsulated \underline{V}ideo \underline{S}ynthesizer that composes T2I and T2V models to enhance both visual fidelity and motion smoothness of generated videos. Our approach utilizes a well-trained diffusion-based T2I model to refine low-quality video frames by treating them as out-of-distribution samples, effectively optimizing them with noising and denoising steps. Meanwhile, we employ T2V backbones to ensure consistent motion dynamics. By encapsulating the T2V temporal-only prior into the T2I generation process, \emph{EVS} successfully leverages the strengths of both types of models, resulting in videos of improved imaging and motion quality. Experimental results validate the effectiveness of our approach compared to previous approaches.
Our composition process also leads to a significant improvement of 1.6x-4.5x speedup in inference time.~\footnote{Source codes:~\url{https://github.com/Tonniia/EVS}}
\end{abstract}

\vspace{-1em}
\section{Introduction}

\begin{figure}
    \centering
    \includegraphics[width=1.0\linewidth]{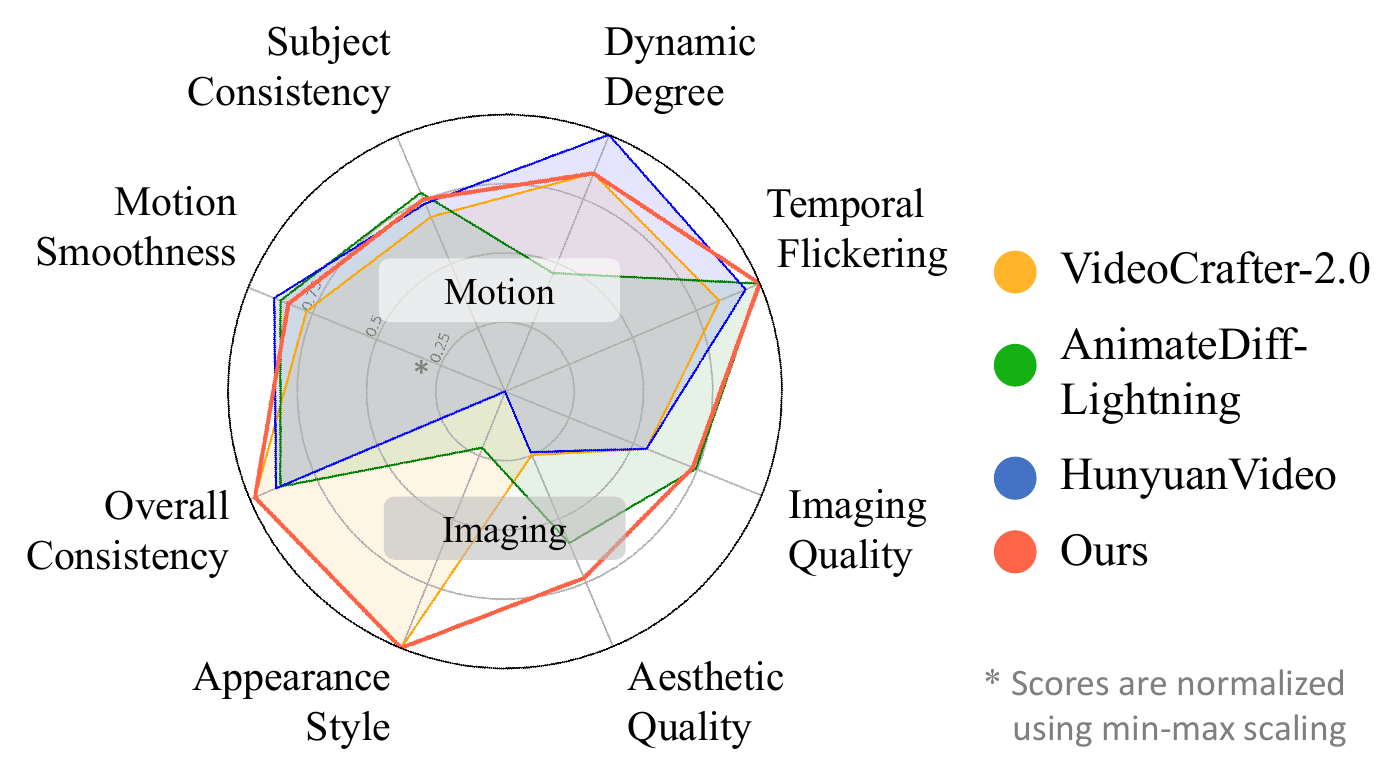}
    \vspace{-2em}
    \caption{Evaluation results of T2I/T2V models on VBench \cite{huang2024vbench}. Results show that motion and imaging qualities are hard to balance in one T2V model. With encapsulated composition of T2I and T2V models, our method effectively combines their advantages.}
    \vspace{-1em}
    \label{fig:vbench_vis}
\end{figure}

Recently, large-scale text-to-video (T2V) models~\cite{pikalab,Gen2,Gen3,sora} have gained significant attention due to their ability to generate realistic videos from textual descriptions. These models leverage vast datasets of text-video pairs, allowing them to learn complex relationships between textual inputs and visual outputs.
Currently, the prominent T2V generation research in the open-source community can largely be categorized into two main approaches. The first approach focuses on training a general T2V diffusion model. This is achieved either by initializing certain modules with pre-trained text-to-image (T2I) models and introduces additional blocks to concentrate on learning the temporal dynamics of videos~\cite{guo2023animatediff,videocrafter2,lavie}, or by training from scratch jointly on images and videos~\cite{kong2024hunyuanvideo,yang2024cogvideox}. In contrast, alternative methods employ T2I models for video synthesis without extensive re-training~\cite{khachatryan2023text2video,qi2023fatezero,zhang2023controlvideo}, which inflate T2I along the temporal axis (i.e., replacing self-attention layers in U-Net with cross-frame attention layers) and successfully maintain the imaging quality of generated videos at the T2I levels. 
Despite these advancements, as shown in Figure~\ref{fig:vbench_vis}, current popular T2V models often struggle to simultaneously ensure high imaging quality and motion quality~\cite{huang2024vbench}, 
which are essential challenges that need to be addressed to improve overall performance of video synthesis.

A straightforward approach to addressing the challenges is to improve the imaging quality of videos generated in the vanilla T2V pipeline by combining T2I models. Yet, T2I models can only be applied in a frame-independent manner, which may lead to flickers between frames.
Thus, explicit consistency constraints are incorporated into the video synthesis pipeline. For instance, Rerender-A-Video~\cite{yang2023rerender} utilizes optical flow to iteratively warp latent features from the previous frame, aligning them with the current frame.
TokenFlow~\cite{geyer2023TokenFlow} explicitly propagates diffusion features and computes inter-frame feature correspondences using Nearest-Neighbor Field (NNF) search. However, these methods are designed for real-world videos and rely on precise optical flow or NNF estimations on input videos with high motion consistency. When these techniques are directly applied to model-generated videos that may exhibit apparent inconsistencies, inaccuracies in estimation can exacerbate these inconsistencies and introduce artifacts.

We propose~\emph{EVS}, a training-free \underline{E}ncapsulated \underline{V}ideo \underline{S}ynthesizer composing of T2I and T2V models to produce videos with significantly balanced imaging and motion qualities, together with large inference speedup compared with vanilla alternating two models.
Specifically, we treat low-quality image frames as out-of-distribution samples~\cite{meng2021sdedit,nie2022diffusion,podell2023sdxl,zhang2023i2vgen} for the T2I model and devise proper noising and denoising steps to pull them back to the high-quality imaging distribution.
As for the underlying T2V model, we employ publicly available backbones that are capable of producing highly consistent and stable videos to enhance the motion smoothness. In the~\emph{EVS} framework, we encapsulate this T2V temporal-only prior into the T2I generation process, mitigating the adverse effects of poor T2V imaging quality. This can be achieved through Selective Feature Injection (SFI), which incorporates inversion features representing spatial details, while allowing the remaining features to be refined by the temporal prior. Experiments on the authoritative benchmark VBench~\cite{huang2024vbench} demonstrate that~\emph{EVS} integrates the advantages of both two types of models, which outperforms baselines and achieves 1.6x-4.5x speedup in inference time.
The results are also summarized in Figure~\ref{fig:vbench_vis}.
In summary, the key contributions of our paper are as follows:
\begin{itemize}
\item We introduce \emph{EVS}, a training-free framework which enhances the imaging and motion qualities of synthesized videos with versatile T2I and T2V diffusion models.
\item We propose a novel encapsulated injection of T2V module into T2I diffusion processes to achieve complementary advantages of T2V and T2I models. 
\item Experiments show that \emph{EVS} effectively improves the imaging and motion qualities of synthesized videos, and achieves 1.6x-4.5x speedup in inference time.
\end{itemize}

\section{Related Works}

\subsection{Video Diffusion Models}
Current diffusion-based T2V methods~\cite{khachatryan2023text2video,qi2023fatezero,videocrafter2,modelscope,lavie,ho2022video,guo2023animatediff,yang2024cogvideox,kong2024hunyuanvideo,fan2025vchitect,lin2024open} can be categorized into two groups. The first category comprises zero-shot methods that require only a pre-trained T2I model~\cite{khachatryan2023text2video,qi2023fatezero,zhang2023controlvideo,singer2022make}. During inference, these methods utilize cross-frame attention to ensure temporal consistency, which are limited to generating videos with simple dynamics and are unable to handle more complex motions.
To address this limitation, some studies fine-tune the T2I model on a single video, enabling the generation of videos with similar motion patterns. However, this approach tends to overfit the specifics of the single video and lacks generalization to other motions~\cite{wu2023tune,shin2024edit,pkdd2024}. 
The second category involves training a general T2V diffusion model on large-scale video data~\cite{guo2023animatediff,videocrafter2,ho2022video,yang2024cogvideox,kong2024hunyuanvideo}. Due to the scarcity of high-quality video-text data, these methods either initialize the T2V spatial module with a pre-trained T2I model or jointly train using both image and video data.

Current T2V benchmarks, such as VBench~\cite{huang2024vbench}, include evaluation metrics such as frame-wise imaging quality, temporal consistency, and dynamic degree. According to the findings~\cite{huang2024vbench,liu2024evalcrafter}, there is currently no model that excels across all dimensions. VideoCrafter-2.0~\cite{videocrafter2}, an open-source T2V model that achieves a balanced and excellent performance across all dimensions, consistently ranking above average in each. It explores various strategies to enhance the learning of temporal modules and mitigate image degradation. Nonetheless, despite these efforts, the resulting imaging quality still falls short when compared to that of dedicated T2I models. Inspired by recent advances in frame-wise video editing~\cite{qi2023fatezero,geyer2023TokenFlow}, one might contemplate improving the imaging quality of individual frames using sophisticated T2I models. However, processing each frame independently may exacerbate inconsistencies among frames, leading to noticeable flickering.


\subsection{Improving Temporal Consistency of Videos}
Previous works~\cite{yang2023rerender,geyer2023TokenFlow,khachatryan2023text2video,qi2023fatezero,Ceylan_2023_ICCV} have considered constraining temporal consistency at both global and local levels. At the global level, they replace self-attention in U-Net with cross-frame attention to regularize the roughly unified appearance; however, this approach is insufficient for ensuring local detail consistency. Rerender-A-Video~\cite{yang2023rerender} and FRESCO~\cite{yang2024fresco} employs optical flow to warp and fuse the latent features, while TokenFlow~\cite{geyer2023TokenFlow} utilizes NNF to compute inter-frame feature correspondences, propagating reference frame features to others. Their explicit enforcement effectively constrains consistency at local level. Nonetheless, achieving explicit correspondence requires that input videos exhibit highly consistent and simple motion~\cite{Liang_2024_CVPR,hu2023videocontrolnet}, with minimal amplitude changes between frames.

With the rapid development of T2V models, there has been increasing interest in exploring the integration of T2I and T2V models~\cite{Liu_2024_CVPR}. BIVDiff~\cite{shi2024bivdiff} firstly employs a T2I model to process individual video frames, followed by a T2V model for temporal smoothing. Consequently, the overall imaging quality aligns with that of the T2V model. VideoElevator~\cite{zhang2024videoelevator} adopts T2I and T2V denoising steps to enhance temporal consistency and imaging quality simultaneously. They break down each sampling step into T2I and T2V from start to finish. Introducing T2I too early can hinder motion understanding, leading to motionless frames. AnyV2V~\cite{ku2024anyv2v} processes the first frame using T2I model and leverage image-to-video~(I2V)~\cite{zhang2023i2vgen} model to handle the entire video. The final results is limited by current I2V ability, which struggles to process videos with complex motion.

\section{\emph{EVS}: The Proposed Method}
In this section, we leverage both T2I and T2V models to enhance imaging quality and motion smoothness of generated videos, without training another refinement model. We first introduces the generation process of T2I and T2V models. After that, we describe two basic composition approaches of T2I and T2V models, discussing and presenting their limitations. We then delve into the specifics of our~\emph{encapsulated composition}, addressing how it overcomes these drawbacks.

\subsection{Preliminaries and Basic Notations} \label{sec:Preliminaries}

\noindent\textbf{T2I.}
T2I models, exemplified by the Latent Diffusion Model (LDM) \cite{LDM}, generate images based on textual descriptions. LDM comprises a pre-trained autoencoder and a U-Net architecture. The encoder compresses an image $x$ into latent space, yielding $z_0 = \mathcal{E}(x)$, while the decoder $\mathcal{D}$ reconstructs $ z_0 $ back to the pixel space. LDM is trained in latent space by estimating various levels of noise added to $z_0$, with the strength parameterized by $\{\Bar{\alpha_t}\}_{t=1}^T$:
\begin{equation}
    z_{t} = \sqrt{\Bar{\alpha}_{t}} z_{0} + \sqrt{1 - \Bar{\alpha}_{t}} \epsilon,
\label{eq:forward_process}
\end{equation}
where $t=0,\cdots, T-1$ is the timestep and $\epsilon$ is the Gaussian random noise. At the inference stage, we sample $z_{t-1}$ based on $z_{t}$ with DDIM sampling ~\cite{songdenoising}: 
\begin{equation}
    z_{t-1}=\sqrt{\overline\alpha_{t-1}} \underbrace{{z}_{t \rightarrow 0}}_{\text {predicted } z_{0}}+\underbrace{\sqrt{1-\overline\alpha_{t-1}} \epsilon_{\theta}\left(z_{t}, t, c\right)}_{\text {direction pointing to } z_{t-1}},
\label{eq:z_prev}
\vspace{-.5em}
\end{equation}
where $ z_{t \rightarrow 0} $ is the predicted clean latent at timestep $t$:
\begin{equation}
    z_{t \rightarrow 0} = (z_{t} - \sqrt{1 - \overline{\alpha}_{t}} \epsilon_{\theta}(z_{t}, t, c) )/{\sqrt{\overline{\alpha}_{t}}},
\label{eq:z_ori}
\end{equation}
$\epsilon_{\theta}$ denotes the noise prediction diffusion model, and $c$ represents the text embedding. For each T2I denoising step, we obtain both $ z_{t-1} $ via Eq.~(\ref{eq:z_prev}) and $ z_{t \rightarrow 0} $ via Eq.~(\ref{eq:z_ori}).


When we aim to improve the imaging quality using T2I models, 
SDEdit noising-denoising procedure is often leveraged~\cite{nie2022diffusion,meng2021sdedit}. Specifically, a noising timestep $t_\text{I}$ and an ending denoising timestep $t_\text{I}'$ are determined. Noise is applied using Eq.~(\ref{eq:forward_process}) to obain $ z_{t_\text{I}} $, then denoising steps are performed to obtain $ {z}_{t_\text{I}'-1}, {z}_{t_\text{I}' \rightarrow 0} $ with enhanced imaging quality. In the following sections, we focus on the predicted $ {z}_{t_\text{I}' \rightarrow 0} $~(denote as $ {z}_0^\text{I} $), as it serves as a crucial link between T2I and T2V. In short, we re-write the process as:
\begin{equation}
{z}_0^\text{I} = \text{T2I}^{\uparrow}(z_0, t_\text{I}, t_\text{I}').
\label{eq:T2I_SDEdit_simple}
\end{equation}

\noindent\textbf{T2V.}
Recent T2V\cite{videocrafter2,lin2024animatediff} training strategy and inference process are consistent with those of T2I models. Additionally, the dimensionality of latent space is expanded along the temporal axis. VideoCrafter-2.0~\cite{videocrafter2} and AnimateDiff~\cite{guo2023animatediff} employ a frame-wise autoencoder to process video frames, ensuring that the clean distribution of T2V is analogous to that of T2I. Similar to Eq.~(\ref{eq:T2I_SDEdit_simple}), we assign a noising timestep $ t_\text{V} $ and an ending timestep $ t_\text{V}'$ to enhance the temporal consistency across the frames from $ z_0 $ to $ {z}_\text{V} $:  
\begin{equation}
    {z}_0^\text{V} = \text{T2V}^{\uparrow}(z_0, t_\text{V}, t_\text{V}').
\label{eq:T2V_SDEdit_simple}
\end{equation}
Note that noise schedules for T2I and T2V are different. Consequently, except for $ z_T $~(both Gaussian noises) and $ z_0 $~(T2I and T2V models share a common autoencoder), the intermediate timestep latents $ z_t $ are drawn from different distributions for T2I and T2V models. Therefore, a latent $ z_t $ from one model cannot be directly fed into the other model.

\noindent\textbf{DDIM Inversion.}
SDEdit~\cite{meng2021sdedit} is typically employed for conditional image generation, where the condition is implicitly remained under noisy latents. Therefore, finding a balance between maintaining condition fidelity and utilizing the diffusion prior presents a significant challenge. DDIM inversion~\cite{songdenoising} deterministically encodes the latent $z_0$ into noisy latent, which can be used to reconstruct $z_0$ through DDIM sampling. To tackle the accumulated error~\cite{mokady2023null} when classifier-free guidance is applied, \cite{hertz2022prompt,Liu_2024_CVPR, chung2024style} collect convolutional features $f_\text{inv}$ and attention features $Q_\text{inv}, K_\text{inv}, V_\text{inv}$ during the DDIM inversion process. These features are selectively injected to replace original features during denoising process in predefined U-Net layers.


\begin{figure}
    \centering
    \includegraphics[width=\linewidth]{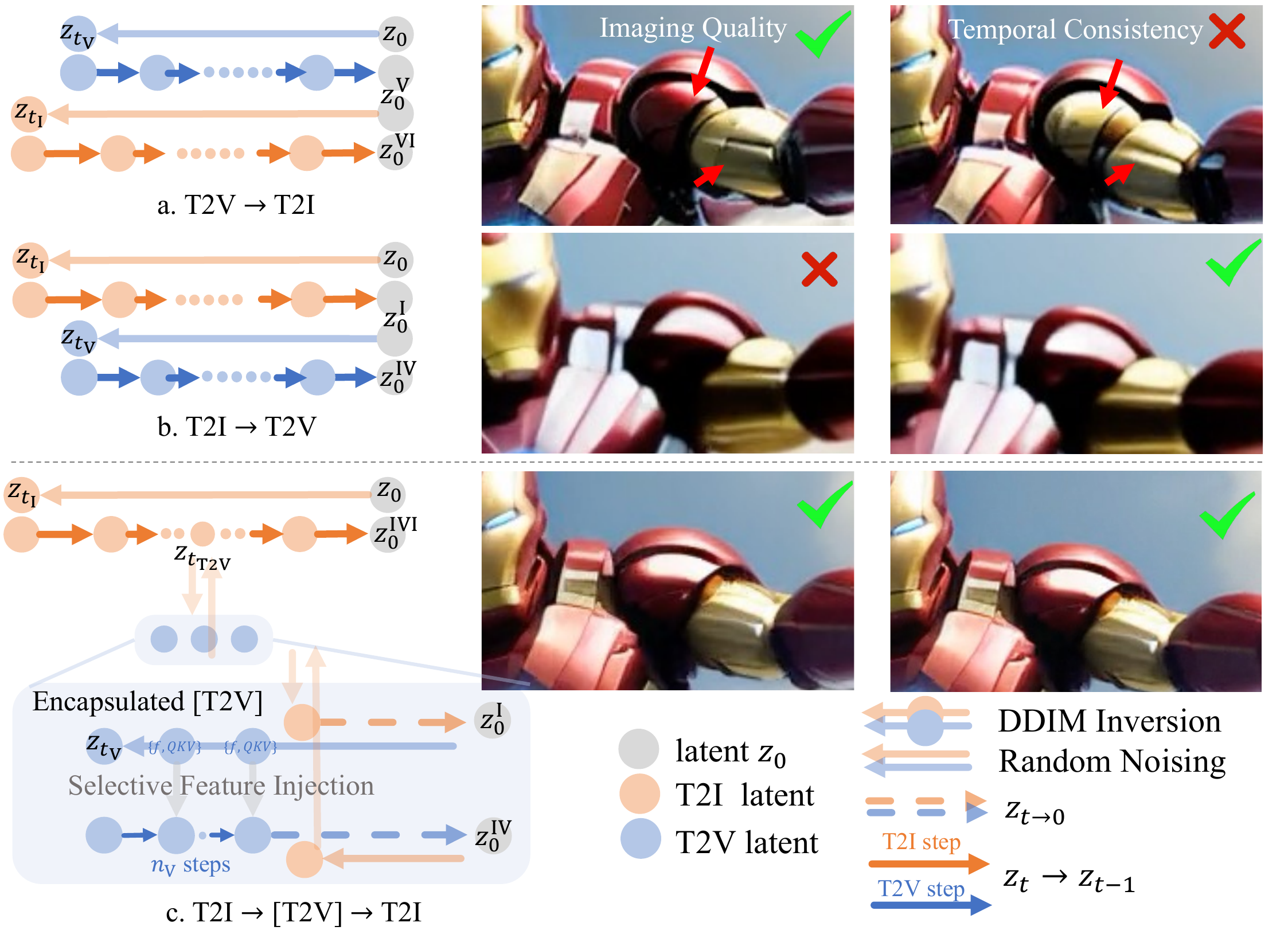}
    \vspace{-1em}
    \caption{\textbf{a/b:} Two basic compositions of T2I and T2V. \textbf{c:} T2I denoising process encapsulated with the [T2V] block.}
    \vspace{-.5em}
    \label{fig:pipeline}
\end{figure}

\subsection{Compositions of T2I and T2V Denoising}\label{sec:pipeline_compose}

In this section, we introduce how to derive the encapsulated composition of T2I and T2V models to generate videos with higher imaging qualities and temporal consistency.

\noindent\textbf{Two Basic Compositions.}
Previous works to compose T2I and T2V models can be summarized into two basic ones. Given a video latent ${z}_0$ with unsatisfied imaging quality and temporal smoothness, the first approach employs the T2V model to produce a smooth video, followed by a few T2I denoising steps to improve the frame imaging quality (see Figure~\ref{fig:pipeline}(a)), similar to previous research on T2I-based video processing~\cite{geyer2023TokenFlow,yang2023rerender}. The noising-denoising process is defined as: 
\begin{equation}
{z}_0^\text{V}  = \text{T2V}^{\uparrow}(z_0, t_\text{V}, 0),\ \  {z}_0^\text{VI}  = \text{T2I}^{\uparrow}({z}_0^\text{V}, t_\text{I}, 0).
\label{eq:T2V_T2I}
\end{equation}
In this scenario, it is trivial to see that frame-wise T2I can re-introduce inconsistencies across frames. The second approach involves starting with the T2I noising and denoising steps, followed by T2V (see Figure~\ref{fig:pipeline}(b))~\cite{shi2024bivdiff}:
\begin{equation}
{z}_0^\text{I}  = \text{T2I}^{\uparrow}(z_0, t_\text{I}, 0),\ \ {z}_0^\text{IV}  = \text{T2V}^{\uparrow}({z}_0^\text{I}, t_\text{V}, 0).
\label{eq:T2I_T2V}
\end{equation}
However, this type of method may lead to a degradation in imaging quality of video frames after refinement via T2V, as the final imaging quality still depends on the T2V model. The challenge arises from the implicit condition of noisy latents in SDEdit, where it is difficult to separate spatial and temporal components. 


We approach this problem from two perspectives. Considering the two basic compositions, the quality of the final video is limited by the model applied at later. Therefore, a natural idea emerges: once we obtain ${z}_0^\text{VI}$ or ${z}_0^\text{IV}$, we can feed them into another round of noising and denoising. This iterative process can introduce a significant amount of redundant steps. To address this, our first strategy is~\emph{Encapsulated Composition}, which efficiently alternates between T2I and T2V stages during the denoising process, thereby eliminating redundancy. To further minimize the impact of the T2V model's disadvantages on imaging quality, we exclusively leverage its temporal prior. This introduces our second strategy of~\emph{leveraging T2V temporal-only prior with Selective Feature Injection} (SFI), where we selectively inject features from DDIM inversion into the denoising process to preserve imaging information from previous T2I steps.

\noindent\textbf{Our Encapsulated Composition.}\label{sec:pipeline_compose}
The key challenge lies in bridging the gap between two latent distributions of T2I and T2V at arbitrary timesteps. Directly utilizing ${z_t}$ from one model (either T2I or T2V) to another is not feasible, particularly when two models employ different sampling methods. Instead of completely denoising the latents using one model and subsequently transitioning to another model~(refer to Eq.~(\ref{eq:T2V_T2I}) and Eq.~(\ref{eq:T2I_T2V})), we propose to leverage the intermediate predicted clean latents to efficiently align the two latent distributions. Specifically, we start with the T2I noising-denoising process:
\begin{equation}
    {z}_0^\text{I}  = \text{T2I}^{\uparrow}(z_0, t_\text{I}, t_\text{T2V}).
\end{equation}

Different from Eq.~(\ref{eq:T2I_T2V}), we do not completely denoise until reaching timestep 0. At timestep $t_\text{T2V}$, the predicted clean latent ${z}_0^\text{I}$ can serve as a shortcut to be connected with T2V, which aligns well with the distribution of clean latent representations in both T2I and T2V models. This allows us to effectively input it into the T2V noising-denoising process:
\begin{equation}
    {z}_0^\text{IV} = \text{T2V}^{\uparrow}({z}_0^\text{I}, t_\text{V}, t_\text{V}-n_\text{V}),
\end{equation}
where $n_\text{V}$ represents how many times the T2V step is executed. This process effectively stabilizes inconsistent video latents $ {z}_0^\text{I} $ from frame-wise T2I denoising into consistent $ {z}_0^\text{IV} $. The $t_\text{V}$ value is independent with T2I shortcut timestep $t_\text{T2V}$, and can therefore be regarded as an \emph{encapsulated block} outside the T2I process. For varying levels of inconsistency in video frames (such as global inconsistency of changing color, or local inconsistency of changing details), different noisy timestep $ t_\text{V} $ will be required. By decoupling this T2V stabilization block from T2I process, we can introduce a significantly larger noise at a later stage for T2I denoising. This approach allows us to explore optimal methods for balancing stabilization with minimal degradation in imaging quality. 
Similarly, $ {z}_0^\text{IV} $ can serve as a shortcut to be connected back to the T2I denoising process for the remaining T2I steps: 
\begin{equation}
    {z}_0^\text{IVI} = \text{T2I}^{\uparrow}({z}_0^\text{IV} , t_\text{T2V}, 0),
\end{equation}
where ${z}_0^\text{IVI}$ is the finally enhanced video latent. Instead of implementing T2V stabilization during every T2I denoising step, we investigate the potential efficiency gained by selecting only one time of applying the block. This aims to conserve computational resources while maintaining effectiveness. Once the T2V block is introduced, the previously improved imaging quality from T2I may deteriorate again, necessitating another round of T2I and T2V processing. This indicates that the final stability of the video frames is significantly influenced by the timing of the last use of the T2V model stabilization, leading to a pipeline represented as T2I+[T2V]+T2I, where the~\emph{encapsulated block} [T2V] needs to be applied only once. 


\noindent\textbf{Leveraging Temporal-Only Prior of T2V.} 
Directly applying T2V-based SDEdit~\cite{meng2021sdedit} simultaneously introduces imaging and temporal prior. Applying DDIM inversion based reconstruction~\cite{songdenoising}, on the other hand, will not introduce any prior. In our work, we wish to maintain imaging quality from T2I steps, and only leverage the temporal prior of T2V. In practice, it is feasible to attain a balanced noising strength with SDEdit. We aim to further minimize the imaging degradation caused by T2V.

\begin{figure}
    \centering
    \includegraphics[width=\linewidth]{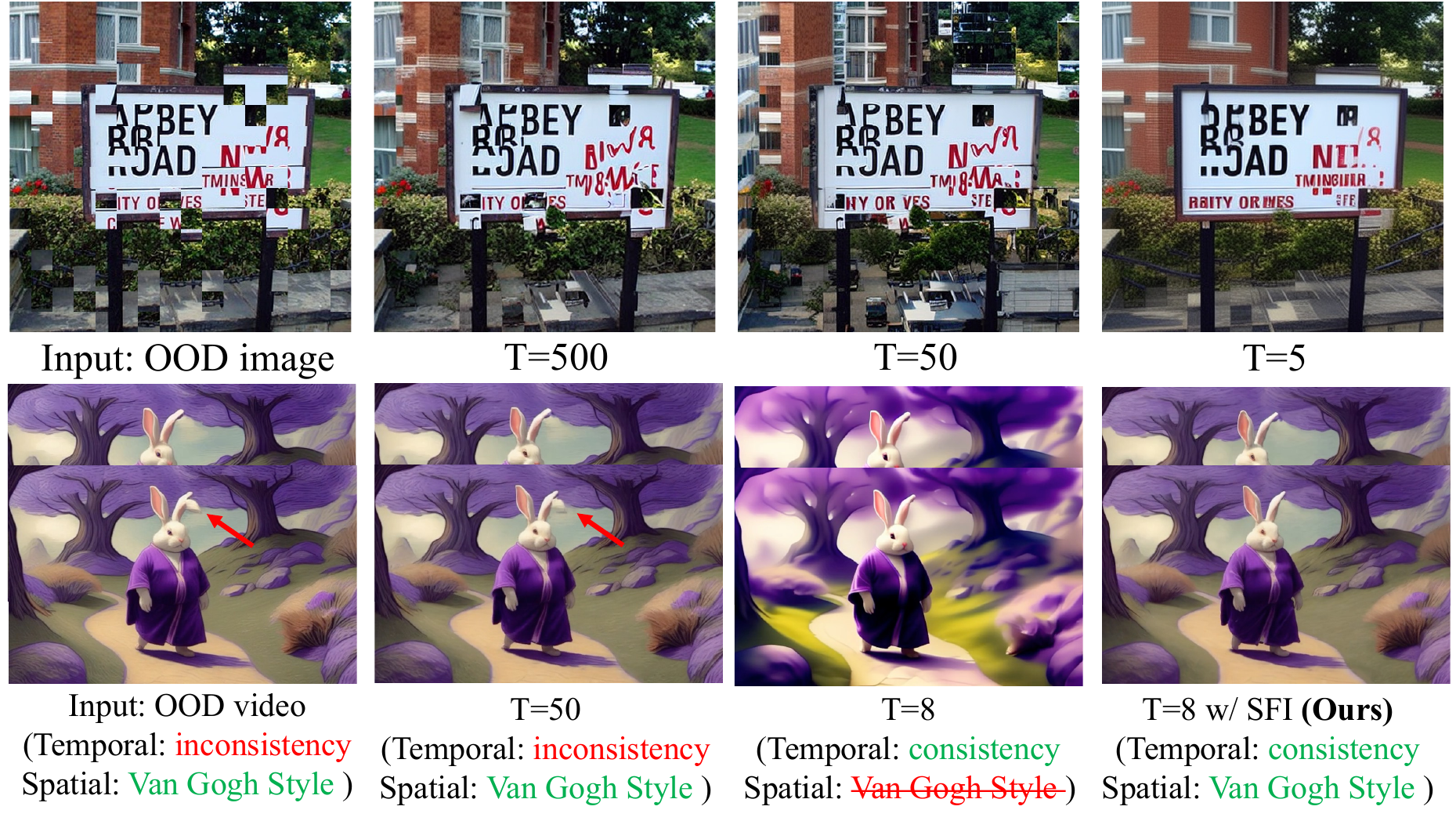}
    \vspace{-1em}
    \caption{\textbf{Row 1:} T2I-based DDIM inversion with different steps. With limited $(T=5)$, T2I prior helps self-rectify images. \textbf{Row 2:} T2V-based DDIM inversion with limited steps~$(T=8)$ simultaneously introduces temporal and spatial prior, but the latter fails to fully capture the original watercolor style. Our Selective Feature Injection~(SFI) strategy can exclusively introduce the temporal prior without imaging style degradation.}
    \vspace{-.5em}
    \label{fig:ddim_sight}
\end{figure}

We start by considering two extreme cases: (1) reconstructing the original video to the maximum extent, using DDIM inversion and full features injection; and (2) leveraging the T2V prior to the maximum extent using DDIM inversion. For a well-trained T2I model, DDIM inversion can reconstruct arbitrary image with enough steps~\cite{songdenoising}, even for out-of-distribution~(OOD) images (as shown in Figure~\ref{fig:ddim_sight} first row, $T=50$). With limited steps~($T=5$), even the reconstruction will fail, it allows for the T2I prior to self-rectify low-quality images. Based on this observation, we regard this partial reconstruction as an opportunity for the T2V model to utilize its prior knowledge. As shown in Figure~\ref{fig:ddim_sight} second row, using the T2V model with fewer steps~($T=8$) can temporally smooth the OOD videos generated from frame-wise T2I-based approaches.



With the above two extreme cases, we can start our probing analysis of U-Net layers to find a satisfied partial reconstruction point, which only leverages the temporal prior of the T2V model. Recent analysis of T2I U-Net decoder layers has shown that, (1) self-attention $\text{Softmax}(QK^\top)$ at deeper layers represent structural information of original images~\cite{liu2024towards,mm2024}. (2) $V$ is closely related to imaging information, i.e, styles and colors, while shallower layers of the U-Net architecture are more related to detailed textures~\cite{chung2024style,tumanyan2023plug}. For our task, we aim at maintaining the imaging information; therefore, we inject $K_\text{inv}, V_\text{inv}$ at predefined layers, while \textit{slightly} perturbing self-attention maps through blending $Q_\text{inv}$ with $Q$ during denoising:
\begin{equation}
    \phi_\text{out} = \text{Attn}(\gamma \cdot Q_\text{inv} + (1-\gamma) \cdot Q, K_\text{inv}, V_\text{inv}),
\end{equation}
where $Q_\text{inv}, K_\text{inv}, V_\text{inv}$ are collected attention features during DDIM inversion process, $Q$ is original query smoothed by T2V during the denoising steps, and $\gamma$ is the blending rate. When $\gamma=1$ and injection is applied in all U-Net layers, the process should result in the upper limit of reconstruction. Then we gradually decrease $\gamma$ and the number of injected layers, progressing from the shallow to the deep layers of the U-Net. This process involves incrementally introducing temporal priors while ensuring the preservation of image quality. Our \textit{Selective Feature Injection} (SFI) strategy simplifies the identification of the balancing point compared to SDEdit, since spatial and temporal information are decoupled in an explainable manner, rather than being implicitly mixed within noisy latent space as in SDEdit.


\begin{algorithm}[ht]
\scriptsize
    \renewcommand{\algorithmicrequire}{\textbf{Input:}}
	\renewcommand{\algorithmicensure}{\textbf{Output:}}
	\caption{The~\emph{EVS} Algorithm}
    \label{SynerT2V_algorithm}
    \begin{algorithmic} 
        \REQUIRE    $z_{0}$: Original video latent;
                    $c$: Text embedding;  
                    $t_\text{T2V}$: [T2V] employ timestep during T2I process;
                    $t_\text{I}, t_\text{V}$: Noising timestep of T2I, T2V; 
                    $n_\text{V}$: [T2V] block denoising steps;
	    \ENSURE ${z}_0^\text{IVI}$: Improved video latent;

        \STATE $z_{t_\text{I}} = \sqrt{\Bar{\alpha}_{t_\text{I}}}z_{0} + \sqrt{1-\Bar{\alpha}_{t_\text{I}}}\epsilon$; 

        \FOR {$t = t_\text{I},t_\text{I}-1,...,1$}
            \STATE $ z_{t \rightarrow 0}, z_{t-1} \leftarrow z_{t}, \epsilon^I_{\theta}(z_{t}, t, c) $; 

            \IF {$t = t_\text{T2V}$}
                
                \STATE $ {z}_0^\text{I} := z_{t_\text{T2V} \rightarrow 0} $; \# bridge to [T2V] 
                \STATE $z_{t_\text{V}}, \{f,QKV\}_\text{inv} = \text{DDIM-inv}({z}_0^\text{I}, t_\text{V})$; 
                \FOR {$t' = t_\text{V},t_\text{V}-1,...,t_\text{V}-n_\text{V}+1$}
                    \STATE $ z_{t' \rightarrow 0}, z_{t'-1} \leftarrow z_{t'}, \epsilon^V_{\theta}(z_{t'}, t, c, \{f,QKV\}_\text{inv}) $; 
                \ENDFOR
                \STATE $ {z}_0^\text{IV}:= {z}_{(t_\text{V}-n_\text{V}) \rightarrow 0}$; \# bridge back to T2I
                \STATE $z_{t_\text{T2V}} = \sqrt{\Bar{\alpha_{t_\text{T2V}}}}{z}_0^\text{IV} + \sqrt{1-\Bar{\alpha}_{t_\text{T2V}}}\epsilon$;
            \ENDIF
        \ENDFOR
        \STATE \textbf{Return} ${z}_0^\text{IVI} := z_{0} $.
    \end{algorithmic}
\end{algorithm}

\subsection{Summary of Our~\emph{EVS} Method}
To summarize, the optimal pipeline in~\emph{EVS} can be characterized as follows: the primary denoising step involves the T2I process for imaging quality enhancement, with intermediate timestep for the application of the [T2V]~\emph{encapsulated block} for temporal motion consistency enhancement. DDIM inversion with \textit{Selective Feature Injection} is \textit{optional} only for challenging cases where the imaging information required and obtained through the T2I model falls entirely outside the T2V domain, making it challenging for SDEdit to obtain a balanced point for both factors. Finally, We present the~\emph{EVS} algorithm pseudo-code in Algorithm 1.



\begin{figure*}[t]
    \centering
    \includegraphics[width=0.95\linewidth]{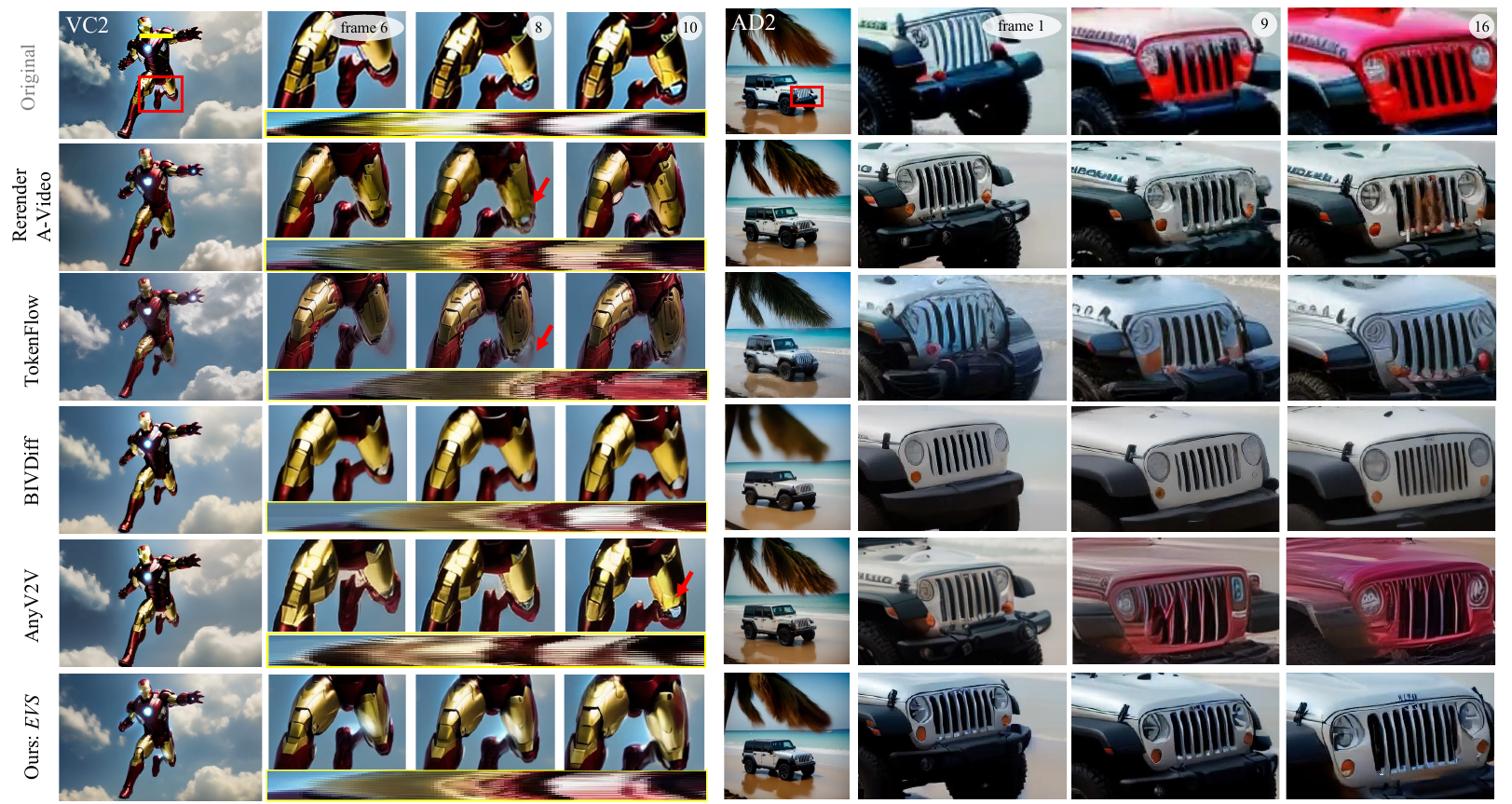}
    \vspace{-.25em}
    \caption{Comparison with baselines. Our method enhances both imaging quality and temporal consistency. Rerender-A-Video and TokenFlow introduce inconsistencies, evidenced by blur and artifacts~(see red frames and flickers in yellow line pixels across frames). T2V temporal smoothing of BIVDiff reverts imaging quality to T2V level, lacking realistic details. AnyV2V is unable to effectively propagate the first frame's enhancement to subsequent frames, resulting in persistent artifacts. We refer the reader to our supplementary material for comprehensive video comparisons with baselines.}
    \vspace{-.25em}
    \label{fig:baselines}
\end{figure*}

\begin{table*}[h]
    \centering
    \resizebox{0.98\textwidth}{!}{
    \begin{tabular}{ccccc|cc||ccccc|cc}
    \hline          & \multicolumn{2}{c}{Consistency~($\uparrow$)} & \multicolumn{2}{c|}{Imaging~($\uparrow$)} & Overall~($\uparrow$) & Time~($\downarrow$) &    & \multicolumn{2}{c}{Consistency~($\uparrow$)} & \multicolumn{2}{c|}{Imaging~($\uparrow$)} & Overall~($\uparrow$) & Time~($\downarrow$) \\
    & MS & SC & DOVER & AP &   &   &   & MS & SC & DOVER & AP &  &\\
    \hline 
    \text{VC2-ori} & 0.9829 & 0.9738 & 55.17 & 4.50 & 0.6435  & - &
    \text{AD2-ori} & \underline{0.9769}  & 0.9484 & 73.32 & 4.69 & 0.5806 & -   \\
    
    \hline 
    \text { Rerender }  & 0.9820 & 0.9745 & \textbf{76.39} & \underline{5.29} & \underline{0.8385}  & 532.15 &
    \text { Rerender }  & 0.9750 & 0.9496 & \textbf{85.05} & \textbf{5.73} & \underline{0.8197}  & 1174.08 \\
    
    
    \text { FRESCO }  & 0.9745 & 0.9706 & \underline{73.59} & 5.44 & 0.7917   & \underline{206.89} &
    \text { FRESCO }  & 0.9667 & 0.9396 & 82.28 & 5.08 & 0.6968   & \underline{494.37} \\
  
    \text { TokenFlow } & 0.9696 & 0.9786 & 64.47 & 4.39 & 0.6759  & 546.94 &
    \text { TokenFlow }  & 0.9630 & \textbf{0.9541} & 72.77 & 4.65 & 0.5369  & 1091.74 \\
    
    \text { BIVDiff }   & \textbf{0.9885}  & \underline{0.9800} & 64.62 & 5.05 & 0.7707  & {352.70} &
    \text { BIVDiff }  & 0.9758 & 0.9513  & 74.39 & 4.91 & 0.6107  & 1073.66 \\
    
    \text { AnyV2V }   & 0.9775  & 0.9540 & 73.10 & 5.11 & 0.7483   & 377.21 &
    \text { AnyV2V }  & 0.9751 & 0.9270  & 76.93 & 4.89 & 0.5998   & {672.71} \\
    \hline 
    
    \text { Ours }  & \underline{0.9881} & \textbf{0.9808} &  {73.20}  & \textbf{5.46} & \textbf{0.8545}  & \textbf{120.86} & 
    \text { Ours } & \textbf{0.9825}  & \underline{0.9530}  &  \underline{84.30}  & \underline{5.42} & \textbf{0.8243}  & \textbf{302.82} \\
    \hline 
    \end{tabular}
    }
    \vspace{-.5em}
    \caption{Quantitative comparison with baselines. First line refers to VC2 and AD2 generated original~(-ori) videos. Our method enhances both imaging quality and consistency and achieves the highest overall score, with 1.6x-4.5x speedup.}
    \label{tab:baselines_vc2}
    \vspace{-.5em}
\end{table*}

\section{Experiments}

\subsection{Experimental Settings}

\noindent\textbf{Dataset.} 
We utilize videos from VBench~\cite{huang2024vbench}, generated by VideoCrafter-2.0~\cite{videocrafter2} \textbf{(VC2)} and AnimateDiff-V2~\cite{guo2023animatediff} \textbf{(AD2)} as our original video dataset. We use the \textit{Overall Consistency} subset\footnote{For each of the 93 prompts provided in the benchmark, we select videos (with id=0) for both VC2 and AD2, totaling 186 videos.}, due to its inclusion of videos characterized by complex movements, and intricate details in the objects. This subset is particularly useful for evaluating temporal motion consistency and imaging quality. The global consistency of videos produced by VideoCrafter-2.0 is well-preserved, particularly regarding the shape and color of subjects. However, the primary issue arises from local inconsistencies, which manifest as flickering and artifacts in details (illustrated in Figure \ref{fig:baselines} Left, where details of Iron Man's legs exhibit noticeable changes across frames). In contrast, AnimateDiff-V2 videos present a more challenging scenario, characterized by global inconsistency (as illustrated in Figure \ref{fig:baselines} Right, where the color of the moving car transitions from white to red). To clearly illustrate the improved imaging quality achieved by T2I models, we resize the videos to twice their original dimensions (\textbf{VC2}: $320 \times 512$, \textbf{AD2}: $512 \times 512$) for input to all baselines.

 
\noindent\textbf{Baselines.} 
We compare our approach with two streams of baselines. Rerender-A-Video~\cite{yang2023rerender}, FRESCO~\cite{yang2024fresco} and TokenFlow~\cite{geyer2023TokenFlow} utilize T2I for frame-wise processing.
TokenFlow also necessitates a precise DDIM inversion and the storage of intermediate features to compute NNF. Since our task involves substantial yet specific adjustments, such as enhancing image quality and adding details, we apply the same inversion strength $s_\text{I} = t_\text{I}/T_\text{I} = 0.4$ for above methods, and our method to ensure a fair comparison. BIVDiff employs a mixed inversion of T2I and T2V inversion to align frame-wise generated latents of T2I with the T2V denoising process, leveraging the T2V model to achieve temporal smoothness of videos. It requires an inversion strength of $s_\text{I} = 1.0$ for mixed inversion in the initial random noise at timestep $t_\text{I} = T_\text{I}$. AnyV2V~\cite{ku2024anyv2v} applys T2I model for first frame processing, followed by the application of the I2V inversion (I2Vgen~\cite{zhang2023i2vgen}) using the same strength as ours. All methods utilize \textit{epiCRealism}\footnote{{https://huggingface.co/emilianJR/epiCRealism}} as T2I, which is specialized in high-quality image synthesis, with total timesteps $T_\text{I}=50$. For BIVDiff and our method, we adopt AnimateDiff-Lightning~\cite{lin2024animatediff} with default $T_\text{V}=8$ as the foundational model for the T2V model due to its superior temporal motion consistency ranking on VBench.

\noindent\textbf{Evaluation Benchmarks.} 
For imaging quality assessment, we adopt \textbf{DOVER}~\cite{wu2023exploring} and the Aesthetic Predictor \textbf{(AP)} V2.5\footnote{{https://github.com/discus0434/aesthetic-predictor-v2-5}}. In terms of motion consistency, we utilize two metrics from VBench~\cite{huang2024vbench}: Motion Smoothness \textbf{(MS)} and Subject Consistency \textbf{(SC)}. The former metric focuses on local consistency by interpolating frames~\cite{li2023amt} $t-1$ and $t+1$ and computing the error with frame $t$. The latter metric emphasizes global consistency by leveraging DINO~\cite{caron2021emerging} feature similarity across all frames. We further calculate an \textbf{Overall} score by averaging the normalized values of the four scores mentioned above. The normalization range is derived from the VBench LeaderBoard\footnote{{https://huggingface.co/spaces/Vchitect/VBench\_Leaderboard}}. 
Additionally, we emphasize time efficiency as we introduce a new composition technique, without time-consuming explicit consistency computing and redundant denoising steps.

\subsection{Comparisons Against Baselines}

Table~\ref{tab:baselines_vc2} shows that our~\emph{EVS} enhances temporal motion consistency and imaging quality of videos generated by VC2 and AD2, achieving an overall better video quality. 
As illustrated in Figure~\ref{fig:baselines} Left, the inaccurate estimation of optical flow in VC2 videos results in significant repainting in Rerender-A-Video. Similarly, the inaccurate estimation of NNF in TokenFlow leads to mismatches between adjacent patches, ultimately resulting in blurring (see mechanical details of Iron Man's legs). In the case of AD2, the inconsistency also exacerbates the issues stemming from inaccuracies in optical flow or NNF estimation. 
As demonstrated in Figure~\ref{fig:baselines} Right, while Rerender-A-Video and TokenFlow may achieve some level of global color consistency, the inaccurate estimation of pixel correspondence continues to induce local inconsistencies (notably seen in the decorative details on the front of the car). This observation is further evidenced in Table~\ref{tab:baselines_vc2}, which indicates that FRESCO encounters a similar issue. The SC metric, which indicates global consistency, shows improvement across all methods. Conversely, the MS metric, reflecting local inconsistency, experiences a decline for most baselines, whereas our approach demonstrates a notable enhancement.
Apart from the overall quality, the inference speed of \emph{EVS} is significantly improved.
Rerender-A-Video and FRESCO requires iterative T2I usage across frames to incorporate optical flow, resulting in a time complexity that scales linearly with frames $N$. 
Tokenflow requires precise T2I DDIM inversion with sufficient number of steps to ensure that the propagation of intermediate features can match the original video content. For two T2V based baselines, BIVDiff requires an inversion strength of $s_\text{I} = 1.0$ for mixed inversion in the initial random noise. AnyV2V relies on precise T2V DDIM inversion to maintain the structural integrity of the source video. Our EVS batchify process all frames without the need for time-consuming accurate inversion. Overall, \emph{EVS} achieves 1.6x-4.5x speedup on these datasets.

\subsection{Ablation Studies}

\noindent\textbf{Compositions Strategy.}
As shown in Table~\ref{tab:pipeline_ablations}, pure T2V or T2I can achieve optimal  consistency or imaging quality. T2I achieves the highest imaging quality score; however,  consistency deteriorates compared original videos due to frame-wise operation. In contrast, T2V produces the highest score in consistency, along with marginally improved imaging quality. 
Two basic compositions, T2I+T2V/T2V+T2I, can slightly balance two aspects, but their results still tends to favor the model used later. As shown in Table~\ref{tab:pipeline_ablations}, T2I+T2V significantly enhances consistency compared to T2I, but results in a notable decline in imaging quality. Conversely, T2V+T2I achieves the highest imaging quality at the expense of poorer consistency. T2I+[T2V]+T2I achieves a balance with comparable consistency from T2V and imaging quality from T2I.


\begin{table}[h]
\vspace{-0.2em}
    \centering
    \resizebox{0.48\textwidth}{!}{
    \begin{tabular}{ccccc|cc}
    \hline    & \multicolumn{2}{c}{Consistency~($\uparrow$)} & \multicolumn{2}{c|}{Imaging~($\uparrow$)} & Overall~($\uparrow$) &  Time~($\downarrow$)\\
    & MS & SC & DOVER & AP &   & \\
    
    \hline 
    \text {VC2-ori}         & 0.9829 & 0.9738 & 55.17 & 4.50 & 0.6435 & - \\
    
    \hline 
    \text { T2I }           & 0.9666  & 0.9710 & 71.96  & \textbf{5.48} & 0.7567 & 108.62 \\
    \text { T2V }           & \textbf{0.9903}  & \textbf{0.9817} & 62.83   & 5.03 & 0.7655 & 60.03 \\
    \text { T2I+T2V }       & \underline{0.9885}  & 0.9800 & 64.62  & 5.05 & 0.7707 & 155.07  \\
    \text { T2V+T2I }       & 0.9868 & 0.9796 & \textbf{75.27} & 5.44 & \underline{0.8474} & 160.31  \\

    \text {T2I+[T2V]+T2I}   & 0.9881  & \underline{0.9808} & \underline{73.20}  & \underline{5.46} & \textbf{0.8545} & 120.86 \\

    \hline
    \end{tabular}
    }
    \caption{Ablation study of T2I and T2V compositions.}
    \label{tab:pipeline_ablations}
    \vspace{-.25em}
\end{table}

\begin{figure}
    \centering
    \includegraphics[width=.85\linewidth]{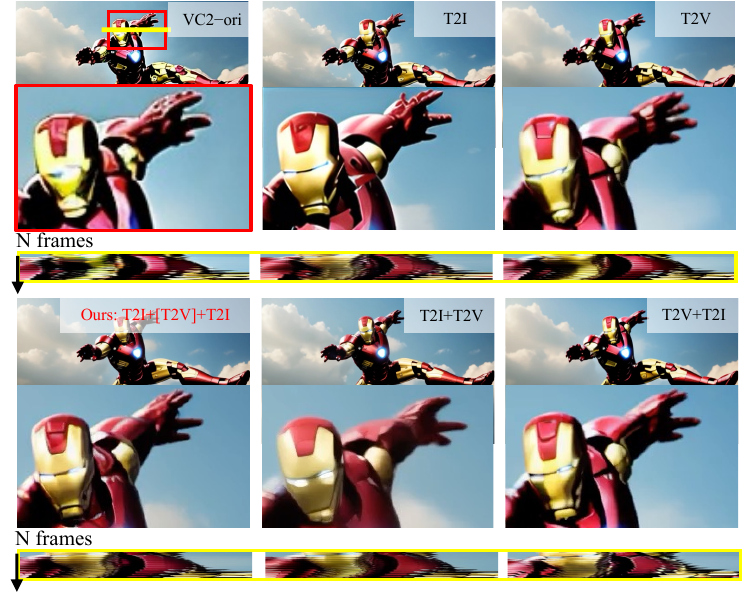}
    \vspace{-1em}
    \caption{T2I+T2V declines to T2V imaging quality~(zoom in to see the blurred face of Iron Man in red frames). T2V+T2I introduces inconsistency similar to T2I~(see flickers in yellow line). Our T2I+[T2V]+T2I balances both aspects.}
    \label{fig:pipeline_ablation}
    \vspace{-.5em}
\end{figure}




    


\begin{figure}
    \centering
    \includegraphics[width=0.86\linewidth]{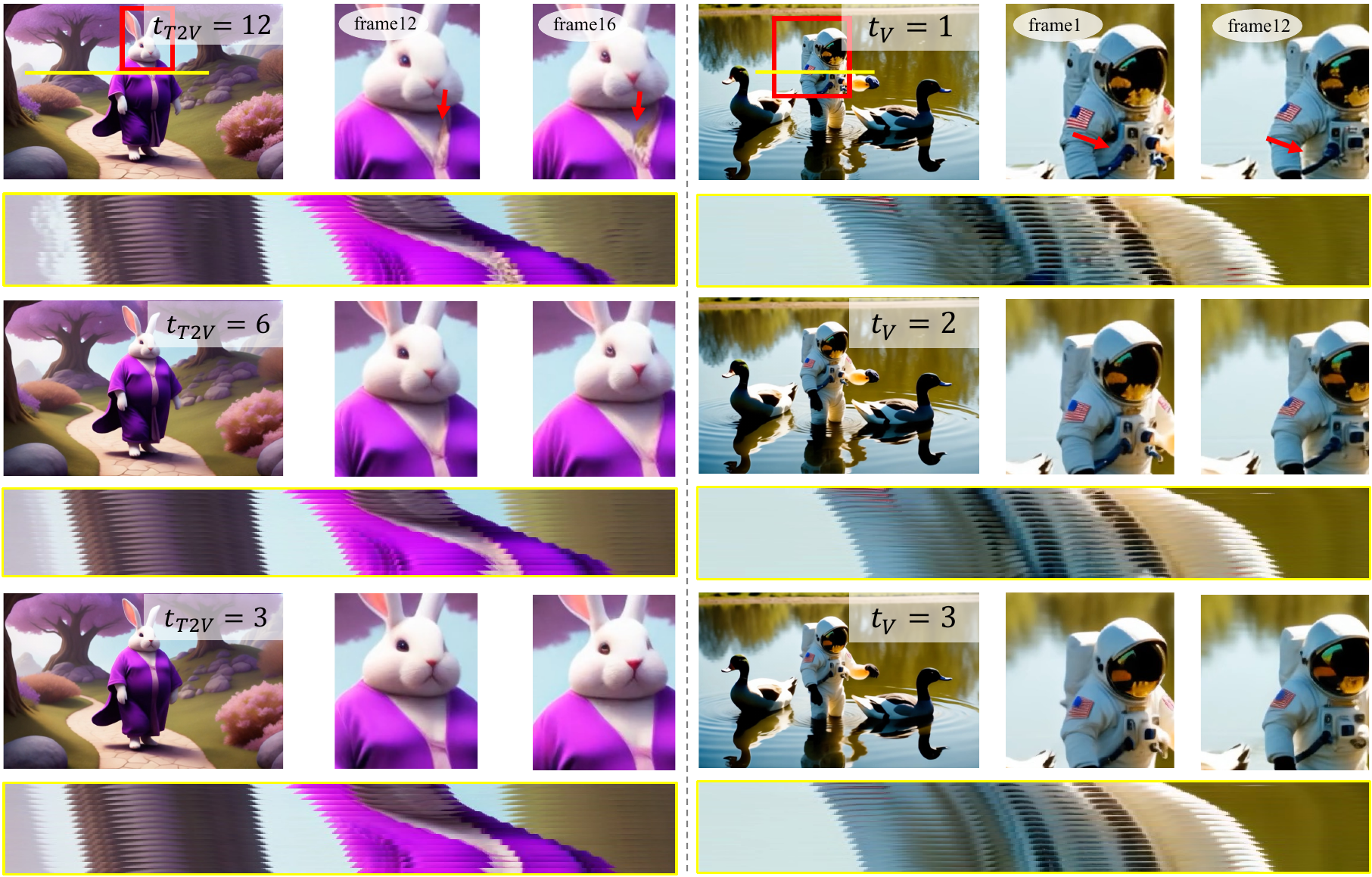}
    \vspace{-.25em}
    \caption{Hyperparameter analysis of the T2V block.}
    \vspace{-1em}
    \label{fig:T2V_ablation}
\end{figure}

\noindent\textbf{Hyperparameter Analysis.}
In Algorithm 1, we utilize four hyperparameters: $t_\text{I}, t_\text{V}, t_\text{T2V}, n_\text{V}$. As illustrated in Figure~\ref{fig:T2V_ablation}, a larger value of $t_\text{T2V}$ (indicating earlier insertion of the T2V block during T2I) allows for more timesteps to be available for T2I, resulting in improved imaging quality but increased inconsistency. A larger $t_\text{V}$ (addition of more noise) can more effectively eliminate inconsistencies. However, excessively large $t_\text{V}$ may cause the frames to converge too closely to the T2V imaging distribution, adversely affecting imaging quality. In the supplementary material, we explore the full combination of these hyperparameters and also apply additional T2V models for enhanced consistency.

\begin{figure}
    \centering
    \includegraphics[width=.90\linewidth]{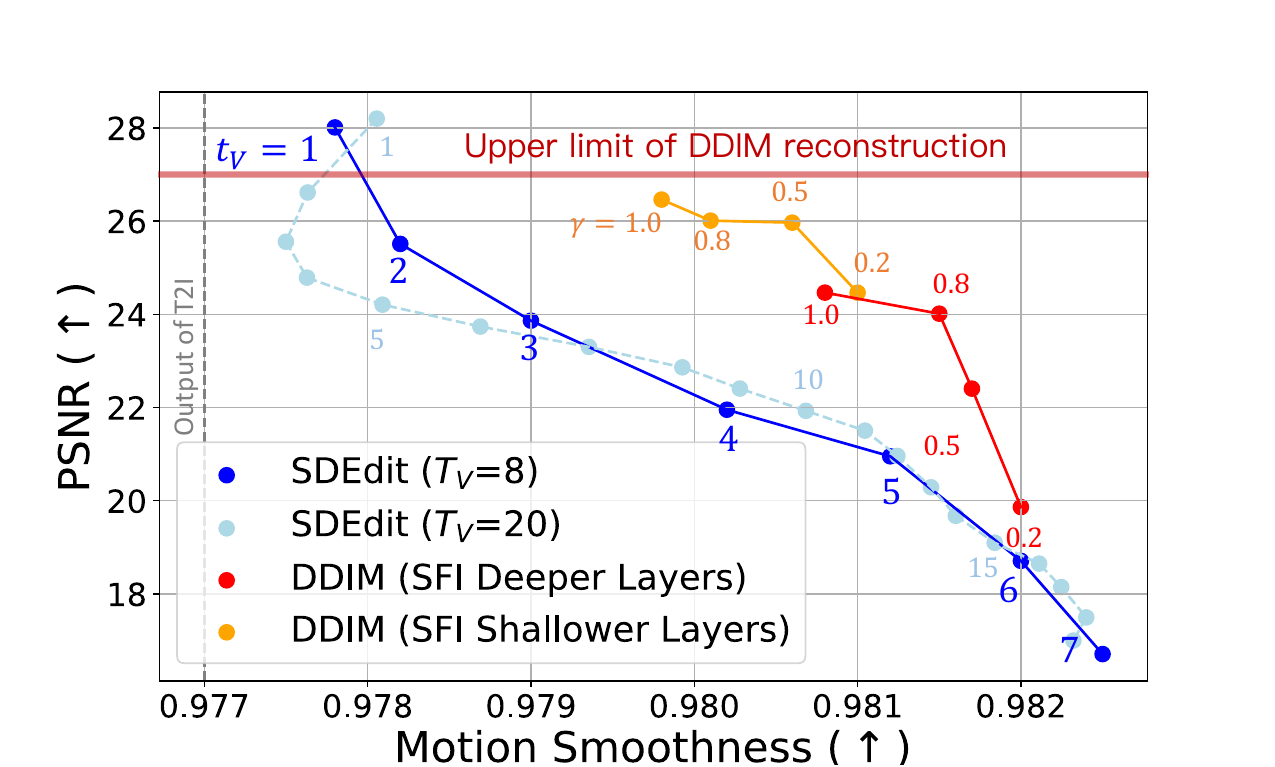}
    \vspace{-.25em}
    \caption{T2V SDEdit v.s. DDIM. It is challenging for SDEdit to strike a balance of maintaining imaging information (PSNR) and introducing T2V motion prior (MS) across all timesteps. DDIM with Selective Feature Injection~(SFI) can find empirical points.}
    \label{fig:ddim_sdedit}
    \vspace{-1em}
\end{figure}

\noindent\textbf{Analysis of SFI.} To validate the effectiveness of the~\emph{Selective Feature Injection} technique in preserving imaging information, we conduct tests on T2I-refined stylized videos from the VBench \textit{Appearance Style} subset. These styles are completely beyond the comprehension of T2V model (e.g. Van Gogh style in Figure~\ref{fig:ddim_sight}). For these videos, we first apply SDEdit with different noising strength. Results in Figure~\ref{fig:ddim_sdedit} indicate that regardless of the total timesteps (8 or 20), as the noising strength increases, there is a noticeable enhancement in temporal smoothness, but this comes at the cost of significant loss in imaging details~(lower PSNR). For DDIM inversion, the overall reconstruction PSNR is notably higher than that of SDEdit under same level of motion smoothness. One can selectively choose injected layers and $Q_\text{inv}$ injected rate $\gamma$: injecting in shallower U-Net layers with higher $\gamma$ can maintain original video in a larger extent. Compared to SDEdit, it is easier to achieve a balanced point. As shown in Figure~\ref{fig:ddim_sdedit}, injecting deeper layers with $\gamma=0.8$ or shallower layers with $\gamma=0.5$ are two optical points. Transitioning from higher $\gamma$ to these two points can enhance motion smoothness with negligible PSNR degradation.

\vspace{-0.4em}
\section{Conclusion}
In conclusion, our novel training-free encapsulated video synthesizer,~\emph{EVS}, successfully bridges the gap between  existing pre-trained T2I and T2V models, resulting in higher-quality video synthesis with enhanced visual fidelity and motion smoothness. It also achieves a significant 1.6x-4.5x speedup in inference time. For our future work, we will continue to improve the video quality for T2V pipelines by refining the T2I/T2V denoising process.

\section*{Acknowledgment}
This work is supported by Key Scientific Research Base for Digital Conservation of Cave Temples (Zhejiang University), State Administration for Cultural Heritage, and Alibaba Research Intern Program.

\small{
\bibliographystyle{ieeenat_fullname}
\bibliography{main}
}


\end{document}